# Task formulation for Extracting Social Determinants of Health from Clinical Narratives

Manabu Torii, Ian M. Finn, Son Doan, Paul Wang, Elly W. Yang, Daniel S. Zisook

**Abstract**

**Objective** The 2022 n2c2 NLP Challenge posed identification of social determinants of health (SDOH) in clinical narratives. We present three systems that we developed for the challenge and discuss the distinctive task formulation used in each of the three systems.

**Materials and Methods** The first system identifies target pieces of information independently using machine learning classifiers. The second system uses a large language model (LLM) to extract complete structured outputs per document. The third system extracts candidate phrases using machine learning and identifies target relations with hand-crafted rules.

**Results** The three systems achieved F1 scores of 0.884, 0.831, and 0.663 in the Subtask A of the Challenge, which are ranked third, seventh, and eighth among the 15 participating teams. The review of the extraction results from our systems reveals characteristics of each approach and those of the SODH extraction task.

**Discussion** Phrases and relations annotated in the task is unique and diverse, not conforming to the conventional event extraction task. These annotations are difficult to model with limited training data. The system that extracts information independently, ignoring the annotated relations, achieves the highest F1 score. Meanwhile, LLM with its versatile capability achieves the high F1 score, while respecting the annotated relations. The rule-based system tackling relation extraction obtains the low F1 score, while it is the most explainable approach.

**Conclusion** The F1 scores of the three systems vary in this challenge setting, but each approach has advantages and disadvantages in a practical application. The selection of the approach depends not only on the F1 score but also on the requirements in the application.

## Background and Significance

Clinical notes are a rich source of information, containing, among others, patient-reported information and clinicians' assessments that are not coded in structured records. Automated extraction and coding of information has been widely studied [1]. Among different types of information sought in clinical notes, social determinants of health (SDOH) have gained attention in the last several years, due to their significance on one's health as well as to their unique availability in clinical notes [2,3]. In the Track-2 of the 2022 n2c2 NLP Challenge [4,5], extraction of SDOH from clinical notes was posed as a shared task, and a corpus annotated with SDOH was prepared by the challenge organizer. The availability of the annotated corpus would further increase interests in this information extraction task and advance the technology toward real-world applications.

This paper focuses on three information extraction systems that we developed for our submissions of the Track-2 in the 2022 n2c2 NLP Challenge, while we defer the background of this Challenge task and the review of related studies to the publications by the Challenge organizers[3–6]. In the corpus prepared for the Challenge, types of annotated phrases are unique and diverse. Relations to be identified among them are difficult to characterize, making the task very different from the conventional event extraction. Each of the three systems we developed employs a different task formulation to tackle this challenge.

## Objective

Natural language processing (NLP) technology has undergone many changes over the years, especially in the last several years [7]. New methods as well as long-standing methods have been evaluated for different clinical NLP tasks in shared-task challenges [8,9]. Besides the performance evaluation results, the task formulation considered for each shared-task challenge has contributed to the clinical NLP field, providing the baselines in designing an information extraction system for the same or related task. Given these backgrounds, two objectives of this paper are as follows:

1. We describe three systems that we developed for the submissions for the 2022 n2c2 NLP Challenge Track-2, which employ both recent and long-standing methods and were ranked high among the participating systems.
2. We present three different formulations of the task that we used in our three systems and discuss the motivation, results, and advantages and disadvantages of each approach.

## Materials and Methods

The Subtask-A of the Track-2, in which we participated, used the Social History Annotation Corpus (SHAC) [3]. The data consisted of 1,316, 188, and 373 clinical narrative texts from MIMIC III [10], which were released respectively as the training, development, and test set. During the challenge period, the training and development sets were made available for the participants to develop systems, and the test set was released for the final evaluation. All these data sets were provided as *brat* annotation files, consisting of narrative text files (.txt) and corresponding annotation files (.ann). Further information of the brat annotation tool can be found in the brat tool paper [11] and on the brat web page [12].

In the SHAC corpus, texts are annotated with *trigger* phrases for five types of SDOH (Alcohol, Tobacco, Drug, Employment, and Living Status) along with their associated *argument* phrases. A subset of the argument phrases, named *labeled arguments*, are normalized to predefined labels (e.g., Status is a labeled argument for Alcohol, normalized to one of the three status values: none, current, or past). The rest of the argument phrases, named *span-only arguments*, do not have labels to normalized to and are "spans only" (e.g., Duration is a span-only argument for Alcohol, annotated for the duration of alcohol use, such as "for eight years"). Further information of the corpus, including annotation examples, can be found in the SHAC corpus paper and in the evaluation guideline document [3,6]. The evaluation script used in the challenge is provided by the organizer on GitHub [13].

During our participation in the challenge, we considered three formulations of the task and implemented three systems as described next. We did not explore the use of additional texts or annotations or the augmentation of the provided data.

*System 1: Sentence classification and sequence labeling*

There are many triggers and arguments in the current task. We observed difficult topics in NLP are involved for their detection (e.g., phrase boundary ambiguity; nested phrase annotations; trigger-argument across sentences; one or more annotated phrases per argument type). However, a good fraction of triggers and arguments look easy to identify (e.g., repeatedly annotated unambiguous phrases). Also, the evaluation metric used in the challenge is forgiving (e.g., phrase spans are not required for the labeled argument). Considering these factors, we convert the given task into a set of simpler tasks that can be tackled by common methods.

In this approach, an input narrative is first split into sentences using a regular expression pattern, and then, two common methods are applied to each sentence, independently:

1. Text classification to identify sentences containing triggers and labeled arguments
2. Sequence labeling to extract triggers and span-only arguments in each sentence.

There are two key observations behind this approach. First, most of the trigger-argument relations are within a single sentence, and there is usually at most one trigger of the same kind within each sentence, which is also noted in the SHAC corpus paper [3]. Second, phrase spans are, in effect, not required in the evaluation of triggers and labeled arguments. That is, labeled arguments are evaluated by the inferred label values only. Triggers are evaluated by the span, but any overlap between the predicted span and the annotated gold span is counted. Therefore, the trigger span is not required in effect if a long enough span is proposed.

The two observations lead to a task formulation that, for triggers and labeled arguments, we only need to classify each sentence whether it implies a particular trigger type or a particular labeled argument, e.g., “Does this sentence report a patient’s alcohol abuse?” or “Does this sentence report a patient’s current alcohol abuse?” As for the span-only argument, the task needs to be tackled as sequence labeling, specifically the common BIO labeling of tokens [14]. A separate model is prepared for each span-only label type and for each trigger type because phrases annotated for span-only arguments and triggers sometimes overlap each other. After triggers and arguments are detected independently, the predictions are merged per sentence. When an argument is predicted by any of the models, the corresponding trigger must be present for it to be reported, and the trigger is additionally predicted, if it is not predicted by the trigger detection methods.

For the text classification, a multi-label text classification model was trained using the Hugging Face Transformers library [15], which is used to make binary classification for 28 targets: 5 triggers and 23 labeled arguments. The implementation follows a publicly available Jupyter notebook example, "Fine-tune BERT for Multi-label Classification" [16]. For the BERT model, `Bio_Discharge_Summary_BERT` was selected [17], because it seemed to yield slightly better performance than the other model we tested, `BioBERT` [18], during the development. For sequence labeling (2), 33 models were trained also using the Hugging Face Transformers library, each of which extracts phrases for a specific trigger and span-only label: 5 triggers and 28 span-only labels. The implementation follows the tutorial "Token classification" in the Hugging Face Course [19]. For the BERT model, between the two models tested, `BioBERT` was used.

A

| .ann Representation | | GPT-J prompt |
|---|---|---|
| E<br>Alcohol: "drinking"<br>Status: "reports"<br>Frequency: "per month"<br>Amount: "2 alcoholic drinks" | A<br>StatusTimeVal: "current" | prompt = '''<br>Make a table about alcohol use in the following story. Use exact words or phrases from the story.<br>She reports drinking 2 alcoholic drinks per month.<br>\| Alcohol \| Amount \| Duration \| Frequency \| History \| Type \| Status \| Inference \|<br>\| drinking \| 2 alcoholic drinks \| unknown \| per month \| unknown \| unknown \| reports \| current \|<br>### |
| E<br>Alcohol: "ETOH"<br>Status: "Denies" | A<br>StatusTimeVal: "none" | Make a table about alcohol use in the following story. Use exact words or phrases from the story.<br>Denies ETOH<br>\| Alcohol \| Amount \| Duration \| Frequency \| History \| Type \| Status \| Inference \|<br>\| ETOH \| unknown \| unknown \| unknown \| unknown \| unknown \| Denies \| none \|<br>### |
| E:<br>Alcohol: "alcoholic drinks"<br>Amount: "Four to five alcoholic drinks"<br>Status: "drinks"<br>Frequency: "per night" | A<br>StatusTimeVal: "current" | Make a table about alcohol use in the following story. Use exact words or phrases from the story.<br>Four to five alcoholic drinks per night.<br>\| Alcohol \| Amount \| Duration \| Frequency \| History \| Type \| Status \| Inference \|<br>\| alcoholic drinks \| Four to five alcoholic drinks \| unknown \| per night \| unknown \| unknown \| drinks \| current \|<br>### |
| E:<br>Alcohol: "Alcohol"<br>Status: "no longer drinking"<br>History: "in 22 months" | A<br>StatusTimeVal: "past" | Make a table about alcohol use in the following story. Use exact words or phrases from the story.<br>Alcohol: Prior history of alcohol abuse, no longer drinking in 22 months per pt, family and PCP.<br>\| Alcohol \| Amount \| Duration \| Frequency \| History \| Type \| Status \| Inference \|<br>\| Alcohol \| unknown \| unknown \| unknown \| in 22 months \| unknown \| no longer drinking \| past \|<br>### |
| E:<br>? | A<br>? | Make a table about alcohol use in the following story. Use exact words or phrases from the story.<br>h/o EtOH abuse but last drink in 2001.<br>\| Alcohol \| Amount \| Duration \| Frequency \| History \| Type \| Status \| Inference \|<br>''' |

B

```
end_sequence='###'

generator_kwargs = {
        'max_new_tokens':100,
        'top_p':1,
        'temperature':.01,
        'clean_up_tokenization_spaces':True,
        'do_sample': True,
        'early_stopping': True,
        'return_full_text': False,
        'pad_token_id': tokenizer.eos_token_id,
        'eos_token_id': int(tokenizer.convert_tokens_to_ids(end_sequence))
}
res = generator(prompt, **generator_kwargs)
print(res)
```

C

Output:

[{'generated_text': '| EtOH | unknown | unknown | unknown | in 2001 | unknown | h/o | past |\n###'}]

Gold:

| EtOH | unknown | unknown | unknown | in 2001 | unknown | h/o | past |

**Figure 1. Few-shotting GPT-J with alcohol narratives**

To few-shot GPT-J for the social history extraction task, we convert the .ann format of the annotated text into a structured table prompt that maintains the essential content but is more compact and amenable for data capture. The word "unknown" is used in all cases where the .ann file does not have an annotation for the given element.

Of note, all information regarding spans is eliminated in the conversion in Figure A. We create a column labeled "Inference" to store categorical annotations. Each E line in the .ann file is translated into a single row in the table prompt, allowing for the possibility of multiple triggers of the same type.

Sample GPT-J generator parameters are shown in Figure B. We use "###" to hint to the model when language generation should cease.

Figure C shows perfect matching of the model output and the gold .ann representation in prompt format.

*System 2: Fine-tuned GPT-J model*

With the general availability of medium and large size language models we were curious to explore how much of the social history information extraction task could be performed by leveraging the knowledge encoded in an LLM as opposed to layering additional knowledge on top. To this end, we attempted to create a system that performed minimal re-representation of the input data in terms of supplementation with linguistic, structural, or clinical context. A toy example of our thought process is shown in Figure 1. Here we illustrate how few-shotting GPT-J with four brief alcohol related narratives allows the model to correctly generate annotations for an unknown example. Figure 1A demonstrates how we sandwich each narrative with a natural language prompt above and a desired table format below. For the few-shot examples we include data from the E lines ("Event" annotation in brat, i.e., a tuple of phrases) and A lines ("Attribute" annotation in brat, i.e., a label value assigned to a phrase) in the brat .ann files [11] but re-formatted to fit the table structure. The rows with gold annotations are placed beneath the header row. We chose the table representation as we suspected it would be more "in distribution" (vs. out of distribution) for GPT-J than other possibilities, including the raw .ann format. In addition, the table format offers flexibility to encompass all E and A information for a given trigger on a single line. For the cases where there are multiple triggers of the same type, we simply add additional rows to the table.

The few-shot example shown in Figure 1A was run with parameters indicated in Figure 1B. The generated text and gold annotation can be seen in Figure 1C. As formulated, the few-shot task is essentially asking the LLM to function as both a question/answer language model (for span extraction) and a few-shot classification model (for categorical assignments). GPT-J performs both tasks admirably in this toy example.

While few-shotting demonstrates the power of models like GPT-J to "learn" from a minimal number of examples, the setup is fragile and does not yield high performance across significant numbers of new inferences. The context window for GPT-J does not permit enough few-shot samples to represent the range of annotations for a given social history element.

Thus, for the actual extraction task we fine-tuned GPT-J using the entirety of the gold .ann files provided.

Fine-tuning was performed on a machine with the following specs: 2 V100 32 Gb graphics cards, Intel Xeon 20 core processor, 11Tb of storage and 512 Gb of RAM. The fine-tuning python script was written in-house but calls Hugging Face's highly abstracted API. DeepSpeed [20] was used to accomplish offloading as follows: stage 1 shards optimizer states across GPUs, stage 2 adds sharding of gradients, stage 3 adds sharding of model parameters and allows offloading of parameters, weights, and optimizer state. Of note, stage 3 allows offloading to NVMe and CPU + memory. While offloading incurs significant I/O burden, it allows for training arbitrarily large models at the expense of memory, CPU compute, and speed.

Just as in the few-shot examples in Figure 1, input to the fine-tuning procedure was provided as single “sandwiches” of natural language prompt, social history narrative, table format, and table rows generated from the annotations in .ann files. Specifically, we incorporated unedited narratives stripped of new lines and span annotations and injected only the “knowledge” that the categorical text is an inference of some sort, and the type of social history data we are looking to generate (in the form of our natural language prompt). In almost every other respect our fine-tuning data is equivalent to using the original files themselves.

While we performed a few experiments with different natural language prompts, we do not have data on the effectiveness of our chosen verbiage “Make a table about **** in the following story. Use exact words or phrases from the story.” Anecdotally the choice of prompt did not seem to impact performance significantly and in this use case, and possibly others, the prompt may have been superfluous. The LLM demonstrated considerable ability to memorize the training data, achieving 93-94% F1 score when applied to the training data.

We did not generate exact spans from GPT-J, hypothesizing that would be challenging. Due to time constraints, we created some simple heuristics to map the model evaluation text back to the narrative string. We did attempt some experiments where we included an additional word on either side of the gold annotation to try and increase specificity in the eventual map back to the narrative. We do not have results on performance from these experiments but anecdotally it seemed to decrease. The loss of information about spans likely resulted in a decreased recall for our effort.

*System 3: NLP pipeline reuse*

In this approach, we regard SDOH information as an event just as the information is annotated and tackle trigger detection, argument detection, and trigger-argument relation extraction. This general framework has been widely used for event extraction [21,22], and subtasks are commonly organized in a pipeline manner, unless they are solved jointly, e.g., System 2. In System 3, we reuse an in-house NLP pipeline built on the UIMA framework [23] to accommodate these subtasks. The pipeline also provides necessary preprocessing, including tokenization, sentence splitting, part-of-speech tagging, and syntactic parsing. The pipeline components integrate different methods and software libraries. For example, for part-of-speech tagging and syntactic parsing, CoreNLP library [24] is used to derive constituent parse trees and dependency graphs.

After preprocessing, an existing pipeline component for phrase detection is applied for the extraction of triggers and argument candidates, where trigger phrases are also assigned with the SDOH type. To this end, a sequence labeling model is trained on trigger and argument phrases annotated in the training corpus using Conditional Random Field (CRF) [25]. Next, a custom component developed for the current task is applied, which links each detected trigger with argument candidates within the same sentence. For linking, hand-crafted rules are implemented, which are based on the constituent span, the dependency link, or any selected text pattern. Rules were developed following the corpus annotation guidelines and provided examples [3,6] and tested on annotations collected from a few notes in the training set.

The existing NLP pipeline, which can provide the system framework and reusable preprocessing components, allowed us to put together this layered system quickly. During our participation in the challenge, however, we could not allocate sufficient time to write rules for many relations and to test them beyond the few examples used the initial development. As reported in the next section, the precision and recall were rather low for this reason. The performance metric reported on this system, therefore, should be interpreted accordingly.

## Results

The three systems were used in our submission of the Subtask-A in the Track-2, where the training, development and test data set were from MIMIC III [10]. Table 2 shows the counts of true positives and predicted positives per target type, obtained on the test data using the evaluation script provided by the challenge organizer. As the table rows show, the evaluation script counts triggers and arguments separately, rather than as trigger-argument pairs or trigger-arguments tuples. Then, it computes the final performance metric from the total counts, which are shown in the first row "OVERALL" in Table 2. The span-only arguments are relatively rare, and the performance metric is mostly based on triggers and labeled arguments. The evaluation results in the two tables show the characteristics of each system as well as that of the evaluation metric.

System 1 tends to predict more triggers and arguments than the other two systems. That would be attributed to multiple models in the system that independently predict targets without considering trigger-argument relations. The current scoring metric favors independent prediction because, as stated above, triggers and arguments are counted separately toward the scoring. For example, if a trigger is predicted correctly, it is counted as one true positive independent of its arguments; if an argument is predicted correctly, the argument and the associated trigger are both counted.

System 2 achieved a good performance metric, and it may be improved further with a larger model and/or larger data. It is notable that this system generates a complete table, where many relations must be considered together, e.g., trigger-argument, argument-argument, and trigger-trigger. The complete relations among triggers and arguments, though restrictive, must help identify consistent answers. For instance, History is a span-only argument type used for a phrase concerning a patient's last use of substance, e.g., "7 years ago." Therefore, it is always related to the Status argument, Status=past, in addition to the trigger, and they should be considered together to report coherent outputs. Among the three systems, this system achieved good or the best results for the three History arguments as in Table 2.

The overall performance of System 3 is not as high as the other two systems in Table 1, but the trigger extraction performance is close to the other two in Table 2. In fact, the baseline performance of trigger extraction is high in this task because there is a relatively small number of recurrent and unambiguous trigger phrases, such as "*ETOH*" (Alcohol), "*IVDU*" (Drug), "*Tobacco*" (Tobacco), "*works*" (Employment), and "*lives*" (Living Status). If a system can memorize those terms, the trigger extraction looks reasonably good. This suggests that the main interest and challenge in this task is argument detection. A major hurdle for System 3 is that there are so many arguments, and it takes time to manually review relations and develop good rules. When a good rule is created, the precision of the extraction can be high, e.g., Drug Status=none or Employment Status=retired. Yet, many rules are needed to boost the recall.

**Table 1.** The evaluation results of our three systems and the first rank system on the Subtask A test set. There are 3,471 annotated instances (positives).

| Subtask | | True Positives | Predicted Positives | Precision | Recall | F1 |
|---|---|---|---|---|---|---|
| A | System 1 | 3,070 | 3,472 | 0.8842 | 0.8845 | 0.8843 |
| | System 2 | 2,776 | 3,210 | 0.8648 | 0.7998 | 0.8310 |
| | System 3 | 2,157 | 3,032 | 0.7114 | 0.6214 | 0.6634 |
| | Rank 1 system | N/A | N/A | 0.9093 | 0.9078 | 0.9008 |
| B | System 1 | 18,376 | 23,261 | 0.7900 | 0.7477 | 0.7683 |
| | Rank 1 system | N/A | N/A | 0.8109 | 0.7703 | 0.7739 |

**Table 2.** The detailed evaluation results of the three systems on the Subtask A test set. In the gold annotation, triggers and labeled arguments are mandatory per "event" and are highlighted in the table.

| SDOH type | argument | subtype | Positives | True Positives | | | Predicted Positives | | |
|---|---|---|---|---|---|---|---|---|---|
| | | | | Sys. 1 | Sys. 2 | Sys. 3 | Sys. 1 | Sys. 2 | Sys. 3 |
| OVERALL | - | - | 3471 | 3070 | 2776 | 2157 | 3472 | 3210 | 3032 |
| Alcohol | Trigger | N/A | 308 | 302 | 288 | 273 | 310 | 307 | 290 |
| | Status | current | 110 | 102 | 87 | 90 | 118 | 108 | 224 |
| | | none | 151 | 144 | 136 | 53 | 148 | 139 | 64 |
| | | past | 47 | 37 | 37 | 0 | 44 | 60 | 2 |
| | Amount | N/A | 47 | 32 | 27 | 15 | 45 | 38 | 35 |
| | Duration | N/A | 6 | 3 | 3 | 0 | 6 | 5 | 7 |
| | Frequency | N/A | 51 | 36 | 29 | 22 | 48 | 49 | 31 |
| | History | N/A | 32 | 14 | 16 | 9 | 28 | 26 | 19 |
| | Type | N/A | 26 | 21 | 16 | 6 | 29 | 23 | 21 |
| Drug | Trigger | N/A | 189 | 182 | 165 | 166 | 190 | 179 | 178 |
| | Status | current | 18 | 12 | 11 | 13 | 19 | 21 | 130 |
| | | none | 153 | 148 | 135 | 47 | 152 | 142 | 48 |
| | | past | 18 | 11 | 10 | 0 | 15 | 16 | 0 |
| | Amount | N/A | 2 | 0 | 0 | 0 | 0 | 4 | 2 |
| | Duration | N/A | 0 | 0 | 0 | 0 | 1 | 1 | 3 |
| | Frequency | N/A | 6 | 1 | 2 | 0 | 4 | 4 | 3 |
| | History | N/A | 10 | 6 | 5 | 1 | 15 | 12 | 7 |
| | Method | N/A | 35 | 20 | 23 | 5 | 23 | 26 | 6 |
| | Type | N/A | 112 | 90 | 89 | 17 | 115 | 110 | 26 |
| Tobacco | Trigger | N/A | 321 | 306 | 283 | 280 | 323 | 302 | 306 |
| | Status | current | 69 | 61 | 44 | 49 | 77 | 61 | 201 |
| | | none | 137 | 129 | 123 | 57 | 135 | 131 | 85 |
| | | past | 115 | 93 | 95 | 10 | 104 | 109 | 20 |
| | Amount | N/A | 105 | 76 | 65 | 47 | 99 | 93 | 71 |
| | Duration | N/A | 51 | 41 | 34 | 32 | 48 | 45 | 40 |
| | Frequency | N/A | 36 | 31 | 25 | 20 | 34 | 34 | 28 |
| | History | N/A | 87 | 57 | 67 | 42 | 83 | 81 | 55 |
| | Method | N/A | 1 | 0 | 0 | 0 | 0 | 0 | 2 |
| | Type | N/A | 20 | 17 | 13 | 4 | 22 | 22 | 14 |
| Employment | Trigger | N/A | 168 | 161 | 113 | 135 | 175 | 122 | 157 |
| | Status | employed | 64 | 57 | 43 | 47 | 67 | 43 | 108 |
| | | homemaker | 1 | 0 | 0 | 0 | 0 | 0 | 0 |
| | | on_disability | 10 | 10 | 2 | 0 | 15 | 5 | 2 |
| | | retired | 38 | 34 | 32 | 33 | 35 | 32 | 35 |
| | | student | 4 | 1 | 0 | 3 | 1 | 0 | 3 |
| | | unemployed | 51 | 47 | 34 | 8 | 54 | 42 | 9 |
| | Duration | N/A | 4 | 1 | 0 | 0 | 3 | 1 | 5 |
| | History | N/A | 6 | 3 | 0 | 2 | 9 | 3 | 6 |
| | Type | N/A | 130 | 91 | 56 | 48 | 129 | 89 | 75 |
| LivingStatus | Trigger | N/A | 242 | 236 | 227 | 228 | 252 | 242 | 243 |
| | Status | current | 234 | 227 | 220 | 220 | 242 | 236 | 241 |
| | | past | 8 | 4 | 5 | 2 | 5 | 5 | 2 |
| | Type | alone | 60 | 59 | 57 | 44 | 62 | 66 | 46 |
| | | homeless | 4 | 4 | 3 | 0 | 5 | 3 | 0 |
| | | with_family | 139 | 136 | 131 | 129 | 143 | 137 | 177 |
| | | with_others | 39 | 25 | 25 | 0 | 34 | 35 | 0 |
| | Duration | N/A | 4 | 1 | 0 | 0 | 5 | 0 | 1 |
| | History | N/A | 2 | 1 | 0 | 0 | 1 | 0 | 2 |

## Discussion

SDOH information in the SHAC corpus is regarded as an “event,” and it is annotated as a trigger with associated arguments. This annotation framework is widely used in event extraction tasks [22]. Meanwhile, it is reported that “[v]arious versions of the event extraction task exist, depending on the goal” [26] and “[t]he definition of an event varies in granularity depending on the desired application of event extraction” [27]. SDOH annotations in the SHAC corpus are particularly unique, in that they are reports on patients’ conditions, rather than event occurrences [2]. Additionally, the annotations include both direct reports (e.g., “past smoker” or “unemployed”) and indirect reports, from which patients’ conditions are inferred (e.g., “He quit smoking” → Smoking Status:past or “former nurse” → Employment Status:unemployed). All these factors make the current task different from the conventional event extraction task.

In the conventional event extraction task, usually, a trigger is a verb, or its nominalization denoting an event occurrence, and arguments are terms syntactically related to the trigger. However, triggers in the SHAC corpus are a mixture of clues indicative of SDOH reports, including section headers (e.g., "Tobacco history: …"), verbs or derivative nouns used to state habits or status (e.g., "smokes" or "smoker"), and any keywords suggestive of SDOH reports (e.g., "cigarettes" or "ppd" (packs per day)). Then, there are four to seven different kinds of arguments for each of the five SDOH targets. Given many kinds of triggers and arguments, relations between them are diverse and complex. Compared to the conventional event extraction task, it is particularly difficult to characterize relations between triggers and arguments.

To mitigate the challenge, System 1 avoided modeling relations and considered independent information extraction tasks. The advantage of this approach is the ease of the complexity in the relation extraction. There are many methods and techniques applicable to the simplified tasks. The disadvantage is that this approach does not extract phrases and relations as in the corpus annotation guidelines. The system is inherently limited, and it cannot extract two triggers in one sentence or trigger-argument across sentences.

System 2 does not simplify the task and generate complete structured outputs. The advantage of this approach is the complete outputs as well as the single end-to-end model dealing with all the relations simultaneously. The disadvantage is that the model behavior cannot be easily understood or modified because it is a single end-to-end model. Also, a large computing resource is needed for LLM, while that can help improve the performance further and can be considered an advantage.

System 3 is based on trigger-argument relation extraction, conforming to the corpus annotation guidelines. The advantage of this approach is the transparency of the extraction procedure and the interpretability of outputs owing to the pipeline architecture and human-readable rules. The disadvantage is that, given many relations in the task, it is time-consuming to analyze them and write good rules. Also, the management of many rules and many pipeline components can be difficult in practice.

As discussed above, the three task formulations have different advantages and disadvantages. Notably, though these systems were evaluated in the same challenge, they are not comparable for building an application. For example, if the goal is to automatically populate a structured database with extracted phrases, System 1, which does not extract trigger phrases and labeled argument phrases, is not applicable. Systems 2 and 3 are applicable to such an application, but the user experience as well as the system maintenance effort is vastly different. If users expect explanation for outputs, a rule-based system like System 2 may be necessary [28]. It must be crucial to understand users' needs and expectation in the application [29].

## Conclusion

In this paper, we describe three information extraction systems that we developed for our participation in the Task-A of the Track-2 in the 2022 n2c2 NLP Challenge, extraction of SDOH from clinical narratives. While the SDOH information is annotated using the event-based annotation framework in the challenge corpus, the meaning of the "trigger" and "argument" is different from the conventional event extraction task. A commonly used approach to event extraction is difficult to apply, due to the diverse and complex relations annotated in this corpus. To overcome this challenge, two alternative task formulations are explored. These approaches have different advantages and disadvantages. The practical utility of the approaches depends on the requirements and expectation in the application.

This paper focuses on SDOH extraction, but the analysis and discussion are applicable to other information extraction tasks in the clinical NLP domain, where target information is often not an "event," but patients' conditions, clinicians' observations and assessments, or various other properties, e.g., severity of a symptom, laterality of an anatomy, a measurement reported for a lab test or a radiographic study, or their combinations. It is desirable if information extraction framework suitable for such targets are investigated further in the clinical NLP domain.

**Acknowledgments**

We thank the organizers and the corpus annotators of the 2022 n2c2 NLP Challenge and the MIMIC project for the data used in the study.